\documentclass{article}

\usepackage{microtype}
\usepackage{graphicx}
\usepackage{subfigure}
\usepackage{booktabs} 

\usepackage{hyperref}

\usepackage{graphicx} 
\usepackage{subfigure}
\usepackage{amsmath}
\usepackage{bm}
\usepackage{multirow}
\usepackage{amssymb}
\usepackage{subfigure}
\usepackage{wrapfig}
\usepackage{float}

\usepackage[accepted]{icml2020}

\icmltitlerunning{A Simple General Approach to Balance Task Difficulty in Multi-Task Learning}

\begin{document}

\twocolumn[
\icmltitle{A Simple General Approach to Balance Task Difficulty in Multi-Task Learning}



\icmlsetsymbol{equal}{*}

\begin{icmlauthorlist}
\icmlauthor{Sicong Liang}{sustech}
\icmlauthor{Yu Zhang}{sustech}
\end{icmlauthorlist}

\icmlaffiliation{sustech}{Department of Computer Science and Engineering, Southern University of Science and Technology, China}

\icmlcorrespondingauthor{Yu Zhang}{yu.zhang.ust@gmail.com}

\icmlkeywords{Machine Learning, ICML}

\vskip 0.3in
]



\printAffiliationsAndNotice

\begin{abstract}

In multi-task learning, difficulty levels of different tasks are varying. There are many works to handle this situation and we classify them into five categories, including the direct sum approach, the weighted sum approach, the maximum approach, the curriculum learning approach, and the multi-objective optimization approach. Those approaches have their own limitations, for example, using manually designed rules to update task weights, non-smooth objective function, and failing to incorporate other functions than training losses. In this paper, to alleviate those limitations, we propose a Balanced Multi-Task Learning (BMTL) framework. Different from existing studies which rely on task weighting, the BMTL framework proposes to transform the training loss of each task to balance difficulty levels among tasks based on an intuitive idea that tasks with larger training losses will receive more attention during the optimization procedure. We analyze the transformation function and derive necessary conditions. The proposed BMTL framework is very simple and it can be combined with most multi-task learning models. Empirical studies show the state-of-the-art performance of the proposed BMTL framework.

\end{abstract}

\newtheorem{theorem}{Theorem}
\newtheorem{lemma}{Lemma}
\newtheorem{definition}{Definition}
\newtheorem{remark}{Remark}
\newtheorem{corollary}{Corollary}

\section{Introduction}

Inspired by human learning ability that human can transfer learning skills among multiple related tasks to help learn each task, multi-task learning \cite{caruana97,zy17b} is to identify common structured knowledge shared by multiple related learning tasks and then share it to all the tasks with the hope of improving the performance of all the tasks. In the past decades, many multi-task learning models, including regularized models \cite{az05,aep06,otj06,jbv08,zy10a,kgs11,ls12,hz16}, Bayesian models \cite{bh03,bcw07,xlck07,zyx10,hh13,hhg15}, and deep learning models \cite{caruana97,msgh16,lqh17,lcwy17,yh17a,yh17b}, have been proposed and those models have achieved great success in many areas such as natural language processing and computer vision.

All the tasks under investigation usually have different difficulty levels. That is, some tasks are easy to learn but other tasks may be more difficult to learn. In most multi-task learning models, tasks are assumed to have the same difficulty level and hence the sum of training losses in all the tasks is minimized. Recently, some studies consider the issue of different difficulty levels among tasks, which exists in many applications, and propose several models to handle this issue. As discussed in the next section, we classify those studies into five categories, including the direct sum approach that includes most multi-task learning models by assuming tasks with the same difficulty level, the weighted sum approach \cite{kgc18,cblr18,ljd19} that learns task weights based on human-designed rules, the maximum approach \cite{mlg12} that minimizes the maximum of training losses of all the tasks, the curriculum learning approach \cite{psl15,lywdlz17,mc17} that learns easy tasks first and then hard tasks, and the multi-objective optimization approach \cite{sk18,lzlzk19} that formulates multi-task learning as a multi-objective optimization problem based on multi-objective gradient descent algorithms.

As discussed in the next section, all the existing studies suffer some limitations. For example, the weighted sum approach relies on human-designed rules to learn task weights, which may be suboptimal to the performance. The maximum approach has a non-smooth objective function, which makes the optimization difficult. Manually designed task selection criteria of the curriculum learning approach are not optimal. It is unclear how to add additional functions such as the regularizer into the multiple objective functions in the multi-objective optimization approach.

In this paper, to alleviate all the limitations of existing studies and to learn tasks with varying difficulty levels, we propose a Balanced Multi-Task Learning (BMTL) framework which can be combined with most multi-task learning models. Different from most studies (e.g., the weighted sum approach, the curriculum learning approach and the multi-objective optimization approach) which minimize the weighted sum of training losses of all the tasks in different ways to learn task weights, the proposed BMTL framework proposes to use a transformation function to transform the training loss of each task and then minimizes the sum of transformed training loss. Based on an intuitive idea that a task should receive more attention during the optimization if the training loss of this task at current estimation of parameters is large, we analyze the necessary conditions of the transformation function and discover some possible families of transformation functions. Moreover, we analyze the generalization bound for the BMTL framework. Extensive experiments show the effectiveness of the proposed BMTL framework.

\section{Preliminaries}
\label{sec:related_work}

In multi-task learning, suppose that there are $m$ learning tasks. The $i$th task is associated with a training dataset denoted by $\mathcal{D}_i=\{(\mathbf{x}^i_j,y^i_j)\}_{j=1}^{n_i}$ where $\mathbf{x}^i_j$ denotes the $j$th data point in the $i$th task, $y^i_j$ is the corresponding label, and $n_i$ denotes the number of data points in the $i$th task. Each data point $\mathbf{x}^i_j$ can be represented by a vector, matrix or tensor, which depends on the application under investigation. When facing classification tasks, each $y^i_j$ is from a discrete space, i.e., $y^i_j\in \{1,\ldots,c\}$ where $c$ denotes the number of classes, and otherwise $y^i_j$ is continuous. The loss function is denoted by $l(f_i(\mathbf{x};\bm{\Theta}),y)$ where $\bm{\Theta}$ includes model parameters for all the tasks and $f_i(\mathbf{x};\bm{\Theta})$ denotes the learning function of the $i$th task parameterized by some parameters in $\bm{\Theta}$. For classification tasks, the loss function can be the cross-entropy function and regression tasks can adopt the square loss. Here the learning model for each task can be any model such as a linear model or a deep neural network with the difference lying in $\bm{\Theta}$. For example, for a linear model where $f_i(\mathbf{x};\bm{\Theta})$ denotes a linear function in terms of $\bm{\Theta}$, $\bm{\Theta}$ can be represented as a matrix with each column as a vector of linear coefficients for the corresponding task. For a deep multi-task neural network with the first several layers shared by all the tasks,  $\bm{\Theta}$ consists of a common part corresponding to weights connecting shared layers and a task-specific part which corresponds to weights connecting non-shared layers.

With the aforementioned notations, the training loss for the $i$th task can be computed as
\begin{equation*}
L(\mathcal{D}_i;\bm{\Theta})=\frac{1}{n_i}\sum_{j=1}^{n_i}l(f_i(\mathbf{x}^i_j;\bm{\Theta}),y^i_j).
\end{equation*}
Difficulty levels of all the tasks are usually different and hence training losses of different tasks to be minimized have different difficulty levels. There are several works to handle the problem of varying difficulty levels among tasks. In the following, we give an overview on those works.

\subsection{The Direct Sum Approach}

The direct sum approach is the simplest and most widely used approach in multi-task learning. It directly minimizes the sum of training losses of all the task as well as other terms such as the regularization on parameters, and a typical objective function in this approach can be formulated as
\begin{equation}
\min_{\bm{\Theta}} \sum_{i=1}^m L(\mathcal{D}_i;\bm{\Theta})+r(\bm{\Theta}),\label{obj_direct_sum}
\end{equation}
where $r(\cdot)$ denotes an additional function on $\bm{\Theta}$ (e.g., the regularization function). In some cases, the first term in problem (\ref{obj_direct_sum}) can be replaced by $\frac{1}{m}\sum_{i=1}^m L(\mathcal{D}_i;\bm{\Theta})$ and it is easy to show that they are equivalent by scaling $r(\cdot)$.

\subsection{The Weighted Sum Approach}

It is intuitive that a more difficult task should attract more attention to minimize its training loss, leading to the weighted sum approach whose objective function is formulated as
\begin{equation}
\min_{\bm{\Theta}} \sum_{i=1}^m w_iL(\mathcal{D}_i;\bm{\Theta})+r(\bm{\Theta}),\label{obj_weighted_sum}
\end{equation}
where $w_i$ is a positive task weight. Compared with problem (\ref{obj_direct_sum}), the only difference lies in the use of $\{w_i\}_{i=1}^m$. When each $w_i$ equals 1, problem (\ref{obj_weighted_sum}) reduces to problem (\ref{obj_direct_sum}).

In this approach, the main issue is how to set $\{w_i\}$. In the early stage, users are required to manually set them but without additional knowledge, users just simply set them to be a identical value, which is just equivalent to the direct sum approach. Then some works \cite{kgc18,ljd19} propose to learn or set them based on data. For example, if using the square loss as the loss function as in \cite{kgc18}, then from the probabilistic perspective such a loss function implies a Gaussian likelihood as $y^i_j\sim\mathcal{N}(f_i(\mathbf{x}^i_j),\sigma_i^2)$, where $\sigma_i$ denotes the standard deviation for the Gaussian likelihood for the $i$th task. Then by viewing $r(\cdot)$ as the negative logarithm of the prior on $\bm{\Theta}$, problem (\ref{obj_weighted_sum}) can be viewed as a maximum a posterior solution of such a probabilistic model, where $w_i$ equals $\frac{1}{\sigma_i^2}$ and $\sigma_i$ can be learned from data. However, this method is only applicable to some specific loss function (i.e., square loss), which limits its application scope. \citet{cblr18} aim to learn task weights to balance gradient norms of different tasks and propose to minimize the absolute difference between the $\ell_2$ norm of the weighted training loss of a task with respect to common parameters and the average of such gradient norms over all tasks scaled by the power of the relative loss ratio of this task. At step $t$, \citet{ljd19} propose a Dynamic Weight Average (DWA) strategy to define $w_i$ as
\begin{equation*}
w_i=\frac{1}{Z}\exp\left\{\frac{L(\mathcal{D}_i;\bm{\Theta}^{(t-1)})}{L(\mathcal{D}_i;\bm{\Theta}^{(t-2)})T}\right\},
\end{equation*}
where $\bm{\Theta}^{(j)}$ denotes the estimation of $\bm{\Theta}$ at step $j$, $Z$ is a normalization factor to ensure $\sum_{i=1}^m w_i=m$, and $T$ is a temperature parameter to control the softness of task weights. Here $w_i$ reflects the relative descending rate. However, manually setting $\{w_i\}$ seems suboptimal.

\subsection{The Maximum Approach}

\citet{mlg12} consider the worst case by minimizing the maximum of all the training losses and formulate the objective function as
\begin{equation}
\min_{\bm{\Theta}} r(\bm{\Theta})+\max_{i} L(\mathcal{D}_i;\bm{\Theta}).\label{obj_minimax}
\end{equation}
To see the connection between this problem and problem (\ref{obj_weighted_sum}) in the weighted sum approach, we can reformulate problem (\ref{obj_minimax}) as
\begin{equation*}
\min_{\bm{\Theta}}\max_{w_i\ge 0\atop\sum_i w_i=1} \sum_{i=1}^m w_iL(\mathcal{D}_i;\bm{\Theta})+r(\bm{\Theta}).
\end{equation*}
According to this reformulation, we can see that the maximum approach shares a similar formulation to the weighted sum approach but $\{w_i\}$ in the maximum approach can be determined automatically. However, the objective function in the maximum approach is non-smooth, which makes the optimization more difficult.

\subsection{The Curriculum Learning Approach}

Curriculum learning \cite{blcw09} and its variant self-paced learning \cite{kpk10}, aim to solve non-convex objective functions by firstly learning from easy data points and then from harder ones. Such idea has been adopted in multi-task learning \cite{psl15,lywdlz17,mc17} by firstly learning from easy tasks and then from harder ones.

In the spirit of curriculum learning, \citet{psl15} take a greedy approach to learn an ordering of tasks where two successive tasks share similar model parameters. However, the analysis in \cite{psl15} is only applicable to linear learners. Built on self-paced learning, \cite{mc17} propose a similar objective function to problem (\ref{obj_weighted_sum}) with $w_i$ defined as
\begin{equation}
w_i\propto \exp\left\{-\frac{1}{T}L(\mathcal{D}_i;\bm{\Theta}^{(t-1)})\right\},\label{equ_CL_rule}
\end{equation}
where $\bm{\Theta}^{(t-1)}$ denotes current estimation of $\bm{\Theta}$ at the previous step and $T$ is a positive hyperparameter. Based on such estimation equation, we can see that a task with a lower training loss in the previous step will have a larger weight at the next step, which follows the philosophy of self-paced learning. Compared with \cite{mc17} which only considers the task difficulty, \citet{lywdlz17} apply self-paced learning to both task and instance levels but it is only applicable to linear models.

\subsection{The Multi-Objective Optimization Approach}

\citet{sk18} and \citet{lzlzk19} study multi-task learning from the perspective of multi-objective optimization where each objective corresponds to minimizing the training loss of a task. Specifically, \citet{sk18} formulate the multi-objective optimization problem as
\begin{equation}
\min_{\bm{\Theta}} \left(L(\mathcal{D}_1;\bm{\Theta}_c,\bm{\Theta}^1_s),\ldots,L(\mathcal{D}_m;\bm{\Theta}_c,\bm{\Theta}^m_s)\right),\label{obj_MGDA}
\end{equation}
where $\bm{\Theta}$ consists of common parameters $\bm{\Theta}_c$ shared by all the tasks and task-specific parameters $\{\bm{\Theta}^i_s\}_{i=1}^m$. One example of such model is the multi-task neural network where $\bm{\Theta}_c$ corresponds to parameters in the first several layers shared by all the tasks and $\bm{\Theta}^i_s$ includes all the parameters in later layers for the $i$th task. In problem (\ref{obj_MGDA}), there are $m$ objectives to be minimized and there is different from aforementioned approaches which have only one objective. The Multi-Gradient Descent Algorithm (MGDA) \cite{desideri12} is used to solve problem (\ref{obj_MGDA}) with respect to $\bm{\Theta}_c$. In each step of MGDA, we need to solve the following quadratic programming problem as
\begin{equation}
\min_{\bm{\alpha}} \left\|\sum_{i=1}^m \alpha_i\mathbf{g}_i\right\|_2^2\quad\mathrm{s.t.}\ \alpha_i\ge 0\ \forall i, \sum_{i=1}^m \alpha_i=1,\label{obj_MGDA_subproblem}
\end{equation}
where $\bm{\alpha}=(\alpha_1,\ldots,\alpha_m)^T$, $\mathbf{g}_i$ denotes the vectorized gradient of $L(\mathcal{D}_i;\bm{\Theta}_c,\bm{\Theta}^i_s)$ with respect to $\bm{\Theta}_c$, and $\|\cdot\|_2$ denotes the $\ell_2$ norm of a vector. After solving problem (\ref{obj_MGDA_subproblem}), we can obtain the optimal $\bm{\alpha}^*$. If $\sum_{i=1}^m \alpha^*_i\mathbf{g}_i$ equals a zero vector, there is no common descent direction for all the tasks and hence MGDA terminates. Otherwise, $\sum_{i=1}^m \alpha^*_i\mathbf{g}_i$ is a descent direction to reduce training losses of all the tasks. In this sense, $\alpha_i$ acts similarly to $w_i$ in the weighted sum approach. However, in this method, the additional function $r(\cdot)$ seems difficult to be incorporated into problem (\ref{obj_MGDA}). Built on \cite{sk18} and decomposition-based multi-objective evolutionary computing, \citet{lzlzk19} decompose problem (\ref{obj_MGDA}) into several subproblems with some preference vectors in the parameter space and then solve all the subproblems. However, preference vectors designed by users seem suboptimal and multiple solutions induced make it difficult to choose which one to conduct the prediction in the testing phase.

%

\section{Balanced Multi-Task Learning}

In this section, we first analyze the limitation of existing works to deal with different levels of task difficulties and then present the proposed BMTL framework.

\subsection{Analysis on Existing Studies}

We first take a look at the learning procedure of the direct sum approach which is fundamental to other approaches. Suppose current estimation of $\bm{\Theta}$ is denoted by $\bm{\Theta}^{(t)}$ and then we wish to update $\bm{\Theta}$ as $\bm{\Theta}^{(t+1)}=\bm{\Theta}^{(t)}+\triangle\bm{\Theta}$. Since $\triangle\bm{\Theta}$ is usually small, based on the first-order Taylor expansion, we can approximate the summed training losses of all the tasks as
\begin{align*}
&\sum_{i=1}^m L(\mathcal{D}_i;\bm{\Theta}^{(t+1)})\\
\approx&\sum_{i=1}^m L(\mathcal{D}_i;\bm{\Theta}^{(t)})+\langle\triangle\bm{\Theta},\nabla_{\bm{\Theta}} L(\mathcal{D}_i;\bm{\Theta}^{(t)})\rangle,
\end{align*}
where $\langle\cdot,\cdot\rangle$ denotes the inner product between two vectors, matrices or tensors with equal size and $\nabla_{\bm{\Theta}} L(\mathcal{D}_i;\bm{\Theta}^{(t)})$ denotes the gradient of $L(\mathcal{D}_i;\bm{\Theta})$ with respect to $\bm{\Theta}$ at $\bm{\Theta}=\bm{\Theta}^{(t)}$. As $\bm{\Theta}$ consists of model parameters of all the tasks, note that some entries in $\nabla_{\bm{\Theta}} L(\mathcal{D}_i;\bm{\Theta}^{(t)})$ will be zero and hence $\nabla_{\bm{\Theta}} L(\mathcal{D}_i;\bm{\Theta}^{(t)})$ is sparse. Then based on problem (\ref{obj_direct_sum}), the objective function for learning $\triangle\bm{\Theta}$ can be formulated as
\begin{equation}
\min_{\triangle\bm{\Theta}} \sum_{i=1}^m \langle\triangle\bm{\Theta},\nabla_{\bm{\Theta}} L(\mathcal{D}_i;\bm{\Theta}^{(t)})\rangle+r(\bm{\Theta}^{(t)}+\triangle\bm{\Theta}).\label{obj_direct_sum_subproblem}
\end{equation}
In problem (\ref{obj_direct_sum_subproblem}), we can see that only the gradient is involved in the learning of $\triangle\bm{\Theta}$. Intuitively, if a task has a large training loss at current step, we hope that at the next step this task should attract more attention to minimize its training loss. So in mathematics, not only the gradient (i.e., $\nabla_{\bm{\Theta}} L(\mathcal{D}_i;\bm{\Theta}^{(t)})$) but also the training loss (i.e., $L(\mathcal{D}_i;\bm{\Theta}^{(t)})$) should be used to learn $\triangle\bm{\Theta}$. However, the direct sum approach cannot satisfy this requirement as revealed in problem (\ref{obj_direct_sum_subproblem}). In the next section, we will see a solution, the proposed BMTL framework, which can satisfy this requirement.

Similar to the direct sum approach, the weighted sum approach with fixed task weights $\{w_i\}_{i=1}^m$ has similar limitations. So the weighted sum approach and other approaches, which take similar formulations to the weighted sum approach with minor differences, propose to use dynamic task weights which depend on model parameters learned in previous step(s). This idea can handle tasks with different difficulty levels to some extent but it brings some other limitations. For example, the weighted sum approach and the curriculum learning approach usually rely on manually designed rules to update task weights, the maximum approach has a non-smooth objective function, and it is unclear to handle additional functions in the multi-objective optimization approach which though has a solid mathematical foundation.

\subsection{The BMTL Framework}

Based on the analysis in the previous section, we hope to use the training losses at current step to learn the update $\triangle\bm{\Theta}$. To achieve this, we propose a BMTL framework as
\begin{equation}
\min_{\bm{\Theta}} \sum_{i=1}^m h(L(\mathcal{D}_i;\bm{\Theta}))+r(\bm{\Theta}),\label{obj_BMTL}
\end{equation}
where $h(\cdot)$ is a function mapping which can transform a nonnegative scalar to another nonnegative scalar. $h(\cdot)$ can be viewed as a transformation function on the training loss and obviously it should be a monotonically increasing function as minimizing $h(L(\mathcal{D}_i;\bm{\Theta}))$ will make $L(\mathcal{D}_i;\bm{\Theta})$ small. For the gradient with respect to $\bm{\Theta}$, we can compute it based on the chain rule as
\begin{align*}
\nabla_{\bm{\Theta}} h(L(\mathcal{D}_i;\bm{\Theta}))=h'(L(\mathcal{D}_i;\bm{\Theta}))\nabla_{\bm{\Theta}} L(\mathcal{D}_i;\bm{\Theta})
\end{align*}
where $h'(\cdot)$ denotes the derivative of $h(\cdot)$ with respect to its input argument. Similar to problem (\ref{obj_direct_sum_subproblem}), the objective function for $\triangle\bm{\Theta}$ is formulated as
\begin{eqnarray}
\min_{\triangle\bm{\Theta}} &&\sum_{i=1}^m h'(L(\mathcal{D}_i;\bm{\Theta}^{(t)}))\langle\triangle\bm{\Theta},\nabla_{\bm{\Theta}} L(\mathcal{D}_i;\bm{\Theta}^{(t)})\rangle\nonumber\\
&&+r(\bm{\Theta}^{(t)}+\triangle\bm{\Theta}).\label{obj_BMTL_subproblem}
\end{eqnarray}
According to problem (\ref{obj_BMTL_subproblem}), $h'(L(\mathcal{D}_i;\bm{\Theta}^{(t)}))$ can be viewed as a weight for the $i$th task. Here $h'(\cdot)$ is required to be monotonically increasing as a larger loss $L(\mathcal{D}_i;\bm{\Theta}^{(t)})$ will require more attention, which corresponds to a larger weight $h'(L(\mathcal{D}_i;\bm{\Theta}^{(t)}))$. In summary, both $h(\cdot)$ and $h'(\cdot)$ are required to be monotonically increasing and they are nonnegative when the input argument is nonnegative. In the following theorem, we prove properties of $h(\cdot)$ based on those requirements.\footnote{All the proofs are put in the appendix.}

\begin{theorem}\label{theorem_h_convexity}
If $h(\cdot)$ satisfies the aforementioned requirements, then $h(\cdot)$ is strongly convex and monotonically increasing on $[0,\infty)$, and it satisfies $h(0)\ge 0$ and $h'(0)\ge 0$.
\end{theorem}

According to Theorem \ref{theorem_h_convexity}, we can easily check whether a function can be used for $h(\cdot)$ in the BMTL framework. It is easy to show that an identity function $h(z)=z$ corresponding to the direct sum approach does not satisfy Theorem \ref{theorem_h_convexity}. Moreover, based on Theorem \ref{theorem_h_convexity}, we can see that compared with the direct sum approach, the introduction of $h(\cdot)$ into problem (\ref{obj_BMTL}) will keep nice computational properties (e.g., convexity) and we have the following results.

\begin{theorem}\label{theorem_problem_convexity}
If the loss function is convex with respect to $\bm{\Theta}$, $\sum_{i=1}^m h(L(\mathcal{D}_i;\bm{\Theta}))$ is convex with respect to $\bm{\Theta}$. If further $r(\bm{\Theta})$ is convex with respect to $\bm{\Theta}$, problem (\ref{obj_BMTL}) is a convex optimization problem.
\end{theorem}

With Theorem \ref{theorem_h_convexity}, the question is how to find an example of $h(\cdot)$ that satisfies Theorem \ref{theorem_h_convexity}. It is not difficult to check that $h(z)=\exp\{\frac{z}{T}\}$ satisfies Theorem \ref{theorem_h_convexity}, where $T$ is a positive hyperparameter. In this paper we use this example to illustrate the BMTL framework and other possible examples such as polynomial functions with nonnegative coefficients that also satisfy Theorem \ref{theorem_h_convexity} will be studied in our future work.

The BMTL framework is applicable to any multi-task learning model no matter wether it is a shallow or deep model and no matter what loss function is used, since $h(\cdot)$ is independent of the model architecture and the loss function. This characteristic makes the BMTL framework easy to implement. Given the implementation of a multi-task learning model, we only need to add an additional line of codes in, for example, the Tensorflow package, to implement $h(\cdot)$ over training losses of different tasks. Hence the BMTL framework can be integrated with any multi-task learning model in a plug-and-play manner.

\subsection{Relation to Existing Studies}
\label{sec:relation_to_existing_sutdies}

When $h(z)=\exp\{\frac{z}{T}\}$, problem (\ref{obj_BMTL}) is low-bounded by problem (\ref{obj_direct_sum}) in the direct sum approach after scaling $r(\cdot)$ plus some constant. To see that, based on a famous inequality that $\exp\{x\}\ge 1+x\ (x\ge 0)$, we have
\begin{align*}
\sum_{i=1}^m h(L(\mathcal{D}_i;\bm{\Theta}))+r(\bm{\Theta})
\ge&\sum_{i=1}^m (1+\frac{1}{T}L(\mathcal{D}_i;\bm{\Theta}))+r(\bm{\Theta})\\
=&m+\frac{1}{T}\sum_{i=1}^m L(\mathcal{D}_i;\bm{\Theta})+r(\bm{\Theta}).
\end{align*}

When $h(z)=\exp\{\frac{z}{T}\}$, problem (\ref{obj_BMTL}) in the BMTL framework is related to problem (\ref{obj_minimax}) in the maximum approach. Based on a well-known inequality that $\max_iz_i\le\ln\left(\sum_{i=1}^m\exp\{z_i\}\right)\le\ln m+\max_iz_i$ for a set of $m$ variables $\{z_i\}_{i=1}^m$, we can obtain the lower and upper bound of $\max_iz_i$ as
\begin{equation*}
\ln\left(\sum_{i=1}^m\exp\{z_i\}\right)-\ln m\le \max_iz_i\le \ln\left(\sum_{i=1}^m\exp\{z_i\}\right).
\end{equation*}
So $\ln\left(\sum_{i=1}^m\exp\{z_i\}\right)$ is closely related to the maximum function and it is usually called the soft maximum function which can replace the maximum function in some case to make the objective function smooth. When replacing the maximum function in problem (\ref{obj_minimax}) with the soft maximum function, it is similar to problem (\ref{obj_BMTL}) in the BMTL framework with an additional logarithm function. Though the soft maximum approach takes a similar formulation to problem (\ref{obj_BMTL}), it does not satisfy Theorem \ref{theorem_h_convexity} and its performance is inferior to the BMTL framework as shown in the next section.




%
%

\subsection{Generalization Bound}

In this section, we analyze the generalization bound for the BMTL framework.


The expected loss for the $i$th task is defined as $R_i(\bm{\Theta})=\mathbb{E}_{(\mathbf{x},y)\sim\mu_i}[l(f_i(\mathbf{x};\bm{\Theta}),y)]$, where $\mu_i$ denotes the underlying distribution to generate the data for the $i$th task and $\mathbb{E}[\cdot]$ defines the expectation. The expected loss for the BMTL framework is defined as $R_h(\bm{\Theta})=\frac{1}{m}\sum_{i=1}^mh(R_i(\bm{\Theta}))$. For simplicity, different tasks are assumed to have the same number of data points, i.e., $n_i$ equals $n_0$ for $i=1,\ldots,m$. It is very easy to extend our analysis to general settings. The empirical loss for the $i$th task is defined as $\hat{R}_i(\bm{\Theta})=\frac{1}{n_0}\sum_{j=1}^{n_0}l(f_i(\mathbf{x}^i_j;\bm{\Theta}),y^i_j)$. The empirical loss for all the tasks is defined as $\hat{R}_h(\bm{\Theta})=\frac{1}{m}\sum_{i=1}^mh(\hat{R}_i(\bm{\Theta}))$. We assume the loss function $l(\cdot,\cdot)$ has values in $[0,1]$ and it is Lipschitz with respect to the first input argument with a Lipschitz constant $\rho$. 

Here we rewrite problem (\ref{obj_BMTL}) into an equivalent formulation as
\begin{equation}
\min_{\bm{\Theta}}\ \sum_{i=1}^m h(L(\mathcal{D}_i;\bm{\Theta}))\quad\mathrm{s.t.}\ r(\bm{\Theta})\le \beta.\label{obj_BMTL_2}
\end{equation}
We define the constraint set on $\bm{\Theta}$ as $\mathcal{C}=\{\bm{\Theta}|r(\bm{\Theta})\le \beta\}$. For problem (\ref{obj_BMTL_2}), we can derive a generalization bound in the following theorem.

\begin{theorem}\label{theorem_general_bound}
When $h(x)=\exp\{x/T\}$, for $\delta>0$, with probability at least $1-\delta$, we have
{\small
\begin{align*}
R_h(\bm{\Theta})\le&\hat{R}_h(\bm{\Theta})+8\rho\nu\mathbb{E}\left[\sup_{\bm{\Theta}\in\mathcal{C}}\left\{\sum_{i=1}^m\frac{\sigma_i}{n_0m}
\sum_{j=1}^{n_0}f_i(\mathbf{x}^i_j;\bm{\Theta})\right\}\right]\\
&+\sqrt{\frac{\eta^2mn_0}{2}\ln\frac{1}{\delta}},
\end{align*}
}\noindent
where $\eta=\frac{2}{m}\exp\{\frac{1}{T}\}(\exp\{\frac{1}{n_0T}\}-1)$ and $\nu=\frac{1}{T}\exp{\frac{1}{T}}$.
\end{theorem}

\begin{remark}
Theorem \ref{theorem_general_bound} provide a general bound for any learning function to upper-bound the expected loss by the empirical loss, the complexity of the learning model reflected in the second term of the right-hand side, and the confidence shown in the last term. Based on $\eta$, the confidence term is $O(\sqrt{\frac{n_0(\exp\{\frac{1}{n_0 T}\}-1)}{m}})$. To see the complexity of the confidence term in terms of $n_0$, according to Lemma \ref{lemma_exponential_function_comparison} in the supplementary material, we have $n_0(\exp\{\frac{1}{n_0 T}\}-1)\ge \frac{1}{T}\exp\{\frac{1}{2n_0T}\}$, implying that the confidence term is $\Theta(\frac{1}{\sqrt{m}}\exp\{\frac{1}{4n_0T}\})$.
\end{remark}

We also consider the case where $h(\cdot)$ is an identity function, i.e., $R_I(\bm{\Theta})=\frac{1}{m}\sum_{i=1}^m R_i(\bm{\Theta})$ which is studied in the direct sum approach. For $R_I(\bm{\Theta})$, we have the following result.

\begin{theorem}\label{theorem_bound_identity_function}
When $h(x)=\exp\{x/T\}$, for $\delta>0$, with probability at least $1-\delta$, we have
{\small
\begin{align*}
R_I(\bm{\Theta})\le&
T\ln\left(8\rho\nu\mathbb{E}\left[\sup_{\bm{\Theta}\in\mathcal{C}}\left\{\sum_{i=1}^m\frac{\sigma_i}{n_0m}
\sum_{j=1}^{n_0}f_i(\mathbf{x}^i_j;\bm{\Theta})\right\}\right]\right.\\
&\left.\qquad+\hat{R}_h(\bm{\Theta})+\sqrt{\frac{\eta^2mn_0}{2}\ln\frac{1}{\delta}}\right).
\end{align*}
}\noindent
\end{theorem}

In Theorem \ref{theorem_bound_identity_function}, it is interesting to upper-bound the the expected loss  $R_I(\bm{\Theta})$ by the empirical loss $\hat{R}_h(\bm{\Theta})$ with a different transformation function.

Based on Theorem \ref{theorem_general_bound}, we can analyze the expected loss for specific models. Due to page limit, a generalization bound for linear models can be found in the supplementary material.

\section{Experiments}

In this section, we conduct empirical studies to test the proposed BMTL framework.

\subsection{Experimental Settings}

\subsubsection{Datasets}

We conduct experiments on four benchmark datasets for classification and regression tasks.

\textbf{Office-31} \cite{saenko2010adapting}: The dataset consists of 4,110 images in 31 categories shared by three distinct tasks: Amazon (A) that contains images downloaded from amazon.com, Webcam (W), and DSLR (D), which are images taken by the Web camera and digital SLR camera under different environmental settings.

\textbf{Office-Home} \cite{venkateswara2017deep}: This dataset consists of 15,588 images from 4 different tasks: artistic images (A), clip art (C), product images (P), and real-world images (R). For each task, this dataset contains images of 65 object categories collected in the office and home settings.

\textbf{ImageCLEF}\footnote{\url{http://imageclef.org/2014/adaptation}}: This dataset contains about 2,400 images from 12 common categories shared by four tasks: Caltech-256 (C), ImageNet ILSVRC 2012 (I), Pascal VOC 2012 (P), and Bing (B). There are 50 images in each category and 600 images in each task.

\textbf{SARCOS}\footnote{\url{http://www.gaussianprocess.org/gpml/data/}}: This dataset is a multi-output regression problem for studying the inverse dynamics of 7 SARCOS anthropomorphic robot arms, each of which corresponds to a task, based on 21 features. By following \cite{zy10a}, we treat each output as a task and randomly sample 2000 data points to form a multi-task dataset.

\subsubsection{Baseline Models}

Since most strategies to balance the task difficulty are independent of multi-task learning models which means that these strategies are applicable to almost all the multi-task learning models, baseline models consist of two parts, including multi-task learning methods and balancing strategies.

$\bullet$ \textbf{Multi-task Learning Methods:} Deep multi-task learning methods we use include ($\romannumeral1$) Deep Multi-Task Learning (\textbf{DMTL}) \cite{caruana97,zllt14} which shares the first hidden layer for all the tasks, ($\romannumeral2$) Deep Multi-Task Representation Learning (\textbf{DMTRL}) \cite{yh17a} which has three variants including \textbf{DMTRL\_Tucker}, \textbf{DMTRL\_TT}, and \textbf{DMTRL\_LAF}, ($\romannumeral3$) Trace Norm Regularised Deep Multi-Task Learning (\textbf{TNRMTL}) \cite{yh17b} with three variants as \textbf{TNRMTL\_Tucker}, \textbf{TNRMTL\_TT}, and \textbf{TNRMTL\_LAF}, and ($\romannumeral4$) Multilinear Relationship Networks (\textbf{MRN}) \cite{lcwy17}.

$\bullet$ \textbf{Balancing Strategies:} As reviewed in Section \ref{sec:related_work}, we choose one strategy from each approach to compare. The strategies we compare include ($\romannumeral1$) the \textbf{Direct Sum (\textbf{DS})} approach formulated in problem (\ref{obj_direct_sum}), ($\romannumeral2$) the Dynamic Weight Average (\textbf{DWA}) method \cite{ljd19} in the weighted sum approach, ($\romannumeral3$) the \textbf{Maximum} (\textbf{Max}) approach formulated in problem (\ref{obj_minimax}), ($\romannumeral4$) the \textbf{Soft Maximum} (\textbf{sMAX}) method discussed in Section \ref{sec:relation_to_existing_sutdies} by minimizing $\log ( \sum_{i=1}^m \exp\{L(\mathcal{D}_i;\bm{\Theta})\})$, ($\romannumeral5$) the \textbf{Curriculum Learning} (\textbf{CL}) method \cite{mc17} by using the self-paced task selection in an easy-to-hard ordering as illustrated in Eq. (\ref{equ_CL_rule}), ($\romannumeral6$) the Multi-Gradient Descent Algorithm (\textbf{MGDA}) method in the multi-objective optimization approach as formulated in problem (\ref{obj_MGDA}), and ($\romannumeral7$) the proposed \textbf{Balanced Multi-Task Learning} (\textbf{BMTL}) framework. Note that among the above seven strategies, the MGDA method is only applicable to the DMTL method while other strategies can be applicable to all the multi-task learning methods in comparison.

For image datasets, we use the VGG-19 network \cite{simonyan2014very} pre-trained on the ImageNet dataset \cite{russakovsky2015imagenet} as the feature extractor based on its fc7 layer. After that, all the multi-task learning methods adopt a two-layer fully-connected network (4096 $\times$ 600 $\times$ $\#$classes) with the ReLU activation in the first layer. The first layer is shared by all tasks to learn a common representation and the second layer is for task-specific outputs. The positive hyperparameter $T$ in the proposed BMTL framework is set to 50.

We use the Tensorflow package \cite{abadi2016tensorflow} to implement all the models. For the optimizer, we use Adam \cite{kingma2014adam} with initial learning rate $\eta_{0}=0.02$ and then iteratively changes the learning rate by $\eta_{p}=\frac{\eta_{0}}{1+p}$, where $p$ is the index of iterations. The size of the mini-batch we use is set to 32.


\subsection{Experimental Results}

\begin{figure*}[!htb]
\vskip -0.1in
  \centering
    \subfigure[DMTL]{\includegraphics[width=0.24\textwidth]{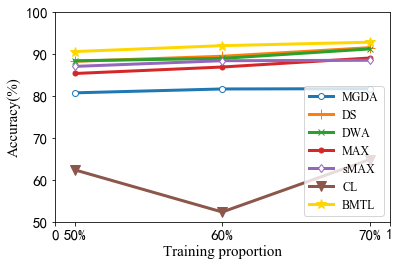}}
    \subfigure[DMTRL\_Tucker]{\includegraphics[width=0.24\textwidth]{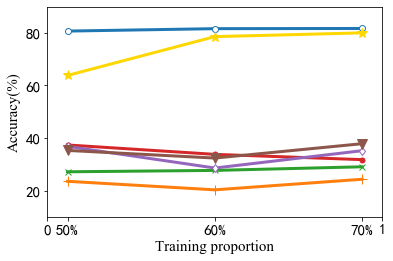}}
    \subfigure[DMTRL\_TT]{\includegraphics[width=0.24\textwidth]{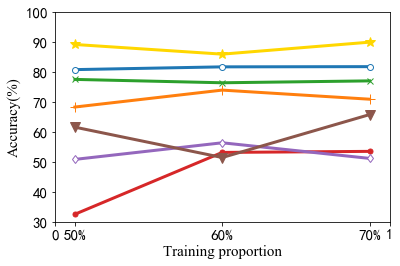}}
    \subfigure[DMTRL\_LAF]{\includegraphics[width=0.24\textwidth]{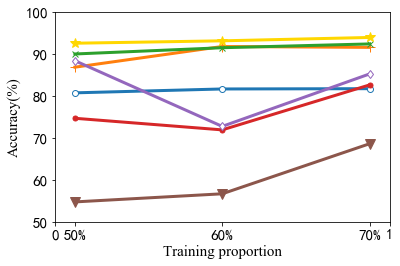}}
    \vskip -0.03in
    \subfigure[TNRMTL\_Tucker]{\includegraphics[width=0.24\textwidth]{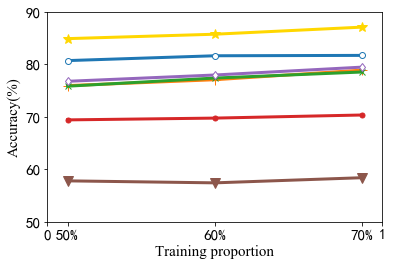}}
    \subfigure[TNRMTL\_TT]{\includegraphics[width=0.24\textwidth]{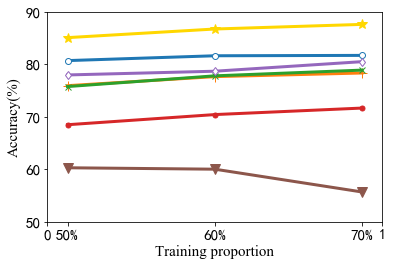}}
    \subfigure[TNRMTL\_LAF]{\includegraphics[width=0.24\textwidth]{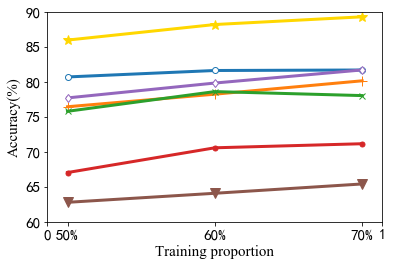}}
    \subfigure[MRN]{\includegraphics[width=0.24\textwidth]{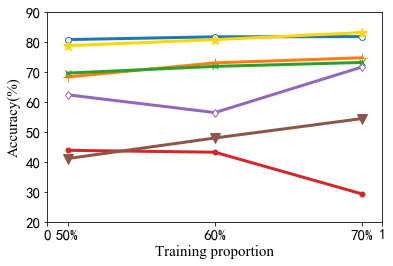}}
  \vskip -0.15in
  \caption{Classification accuracy of different balancing strategies applied to different multi-task learning methods on the \textit{Office-31} dataset by varying the training proportion.}
    \label{fig:office_31}
\vskip -0.1in
\end{figure*}
\begin{figure*}[!htb]
\vskip -0.1in
  \centering
    \subfigure[DMTL]{\includegraphics[width=0.24\textwidth]{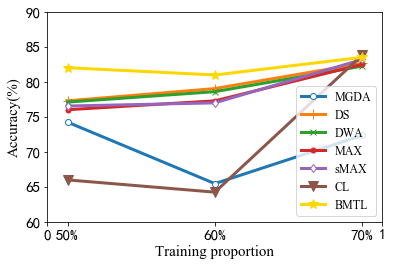}}
    \subfigure[DMTRL\_Tucker]{\includegraphics[width=0.24\textwidth]{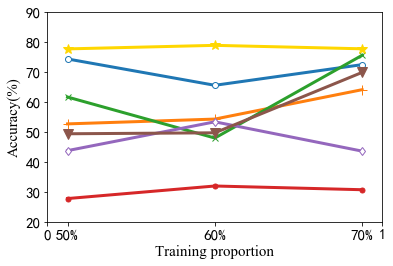}}
    \subfigure[DMTRL\_TT]{\includegraphics[width=0.24\textwidth]{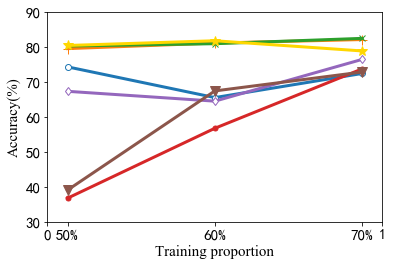}}
    \subfigure[DMTRL\_LAF]{\includegraphics[width=0.24\textwidth]{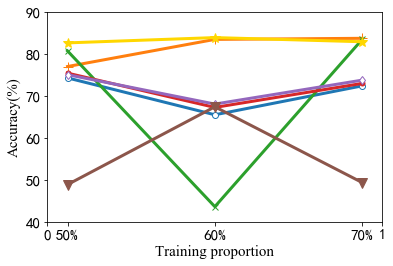}}
    \vskip -0.03in
    \subfigure[TNRMTL\_Tucker]{\includegraphics[width=0.24\textwidth]{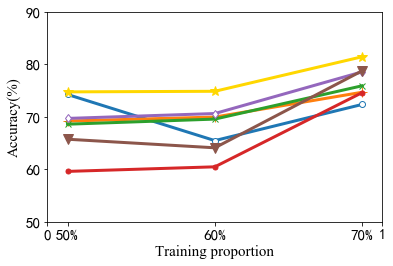}}
    \subfigure[TNRMTL\_TT]{\includegraphics[width=0.24\textwidth]{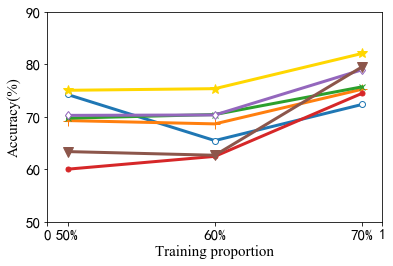}}
    \subfigure[TNRMTL\_LAF]{\includegraphics[width=0.24\textwidth]{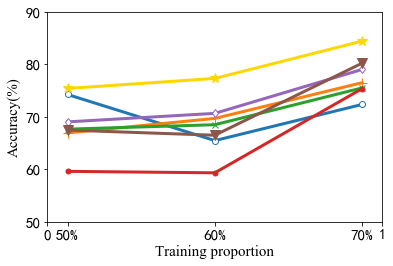}}
    \subfigure[MRN]{\includegraphics[width=0.24\textwidth]{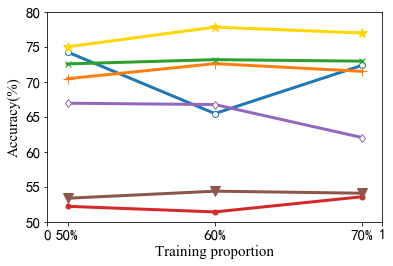}}
  \vskip -0.1in
  \caption{Classification accuracies of different balancing strategies applied to different multi-task learning methods on the \textit{ImageCLEF} dataset by varying the training proportion.}
    \label{fig:imageCLEF}
\vskip -0.1in
\end{figure*}
\begin{figure*}[!htb]
\vskip -0.1in
  \centering
    \subfigure[DMTL]{\includegraphics[width=0.24\textwidth]{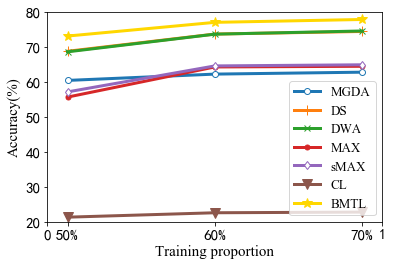}}
    \subfigure[DMTRL\_Tucker]{\includegraphics[width=0.24\textwidth]{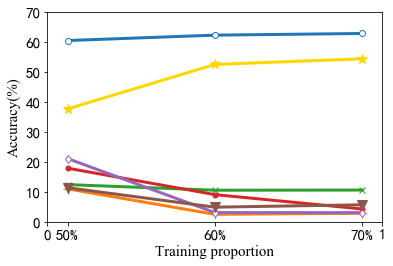}}
    \subfigure[DMTRL\_TT]{\includegraphics[width=0.24\textwidth]{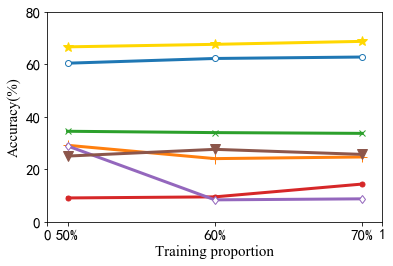}}
    \subfigure[DMTRL\_LAF]{\includegraphics[width=0.24\textwidth]{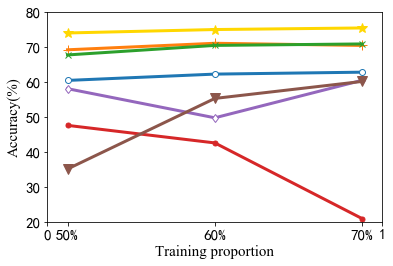}}
    \vskip -0.03in
    \subfigure[TNRMTL\_Tucker]{\includegraphics[width=0.24\textwidth]{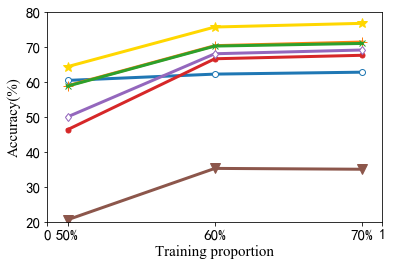}}
    \subfigure[TNRMTL\_TT]{\includegraphics[width=0.24\textwidth]{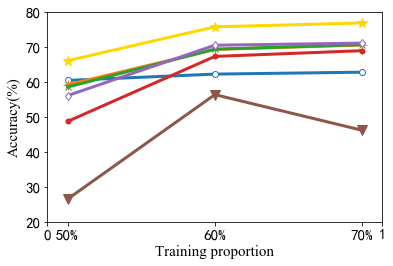}}
    \subfigure[TNRMTL\_LAF]{\includegraphics[width=0.24\textwidth]{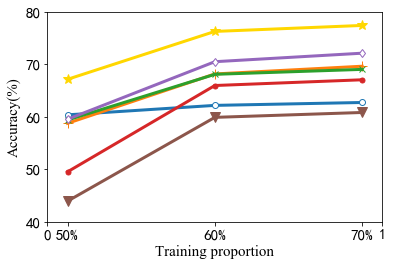}}
    \subfigure[MRN]{\includegraphics[width=0.24\textwidth]{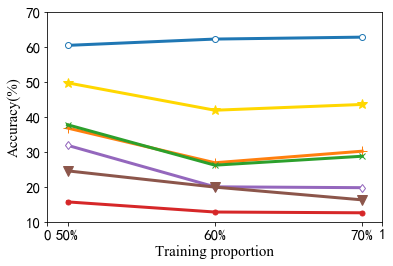}}
  \vskip -0.1in
  \caption{Classification accuracies of different balancing strategies applied to different multi-task learning methods on the \textit{Office-Home} dataset by varying the training proportion.}
    \label{fig:officehome}
\vskip -0.1in
\end{figure*}

To analyze the effect of the training proportion to the performance, we evaluate the classification accuracy of all the methods by using the training proportion as 50\%, 60\%, and 70\%, respectively, and plot the average test accuracy of all the balancing strategies applied to all the multi-task learning methods in Figures \ref{fig:office_31}-\ref{fig:officehome}. Each experimental setting repeats five times and for clear presentation, Figures \ref{fig:office_31}-\ref{fig:officehome} only contain the average accuracies. 

\begin{figure}[!]
\centering
\includegraphics[width=0.4\textwidth]{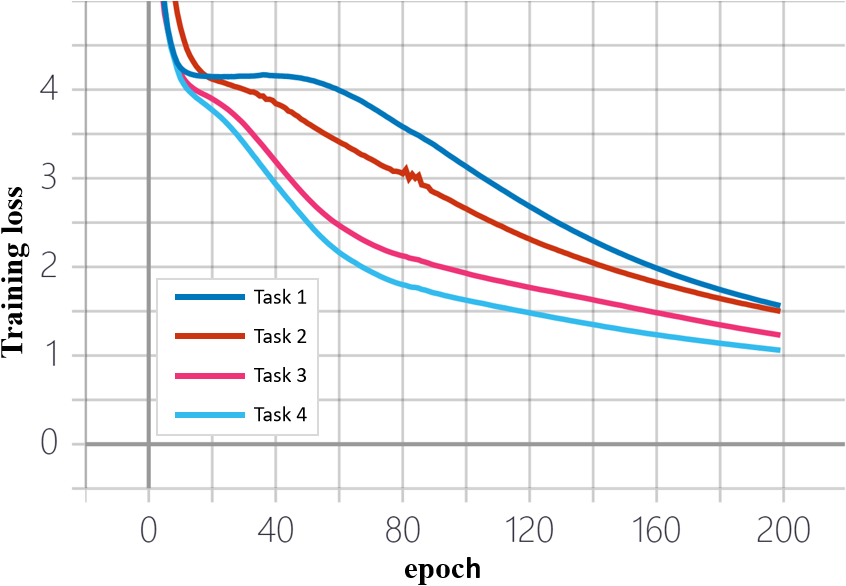}
\caption{Curves of training losses of tasks from the \textit{Office-Home} dataset for the BMTL combined with the DMTRL\_TT method with the training proportion as 0.5.}
\label{fig:training_loss}
\end{figure}

According to Figures \ref{fig:office_31}-\ref{fig:officehome}, we can see that when the training proportion increases, the performance of all the balancing strategies on all the multi-task learning models almost improves with some exceptions due to the sensitivity to the initial values for model parameters in multi-task learning models. Moreover, we observe that compared with all the balancing strategies, the proposed BMTL framework improves every multi-task baseline method with different training proportions, which proves the effectiveness and robustness of the BMTL framework.

From the results shown in Figure \ref{fig:office_31}(b), \ref{fig:officehome}(b) and \ref{fig:officehome}(h), we can see that the MGDA method outperforms other losses balancing strategies that are based on the DMTRL\_Tucker and MRN methods. One reason is that the MGDA method is specific to the DMTL method and inapplicable to other DMTL methods and hence the comparison here is not so fair. In those settings, the proposed BMTL framework still significantly boosts the performance of the DMTRL\_Tucker and MRN methods. 

\begin{figure*}[!htb]
\vskip -0.1in
  \centering
    \subfigure[The legend of figure]{\includegraphics[width=0.3\textwidth]{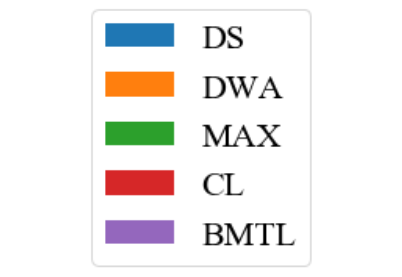}}
    \subfigure[DMTL]{\includegraphics[width=0.3\textwidth]{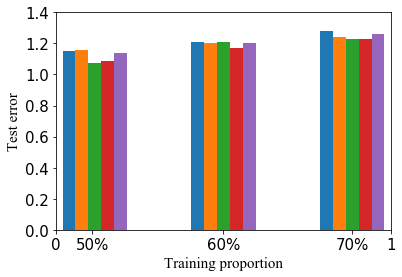}}
    \subfigure[DMTRL]{\includegraphics[width=0.3\textwidth]{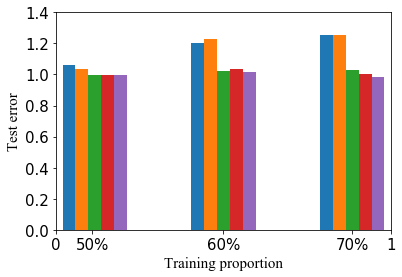}}
    \vskip -0.03in
    \subfigure[TNRMTL\_Tucker]{\includegraphics[width=0.3\textwidth]{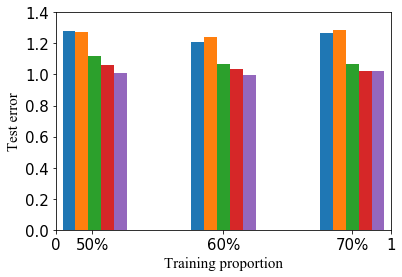}}
    \subfigure[TNRMTL\_TT]{\includegraphics[width=0.3\textwidth]{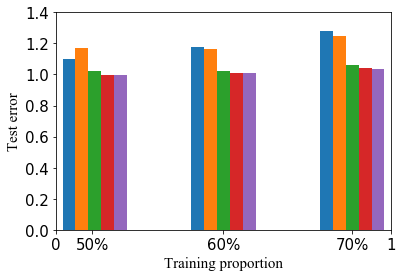}}
    \subfigure[TNRMTL\_LAF]{\includegraphics[width=0.3\textwidth]{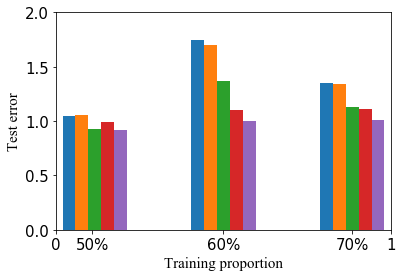}}
\vskip -0.15in
  \caption{Regression errors when varying the training proportion on the \textit{SARCOS} dataset.}
    \label{fig:sarcos}
    \vskip -0.1in
\end{figure*}

For the SARCOS dataset, we use the mean square error as the performance measure. The results are shown in Figure \ref{fig:sarcos}. As shown in Figure \ref{fig:sarcos}, the proposed BMTL framework outperforms other balancing strategies, especially based on the TNRMTL methods, which demonstrates the effectiveness of the BMTL framework in this dataset.

\subsubsection{Analysis on Training Losses}

For the proposed BMTL framework based on the DMTRL\_TT method, we plot the training losses of different tasks from the Office-Home dataset in Figure \ref{fig:training_loss}. From the curves of the training loss, we can observe that tasks with larger training losses draw more attention and decrease faster than other tasks during the training process.

\section{Conclusion}

In this paper, we propose a balanced multi-task learning framework to handle tasks with unequal difficulty levels. The main idea is to minimize the sum of transformed training losses of all the tasks via the transformation function. The role of the transformation function is to make tasks with larger training losses receive larger weights during the optimization procedure. Some properties of the transformation function are analyzed. Empirical studies conducted on real-world datasets demonstrate the effectiveness of the BMTL framework. In our future work, we will investigate other examples of $h(\cdot)$ such as polynomial functions.



\bibliographystyle{icml2020}
\bibliography{BMTL}

\section*{Appendix}

\subsection*{Lemma \ref{lemma_exponential_function_comparison} and Its Proof}

\begin{lemma}\label{lemma_exponential_function_comparison}
For $x>0$, we have
\begin{equation*}
\frac{\exp\{x\}-1}{x}\ge \exp\left\{\frac{x}{2}\right\}.
\end{equation*}
\end{lemma}
{\bf Proof}. Based on the Taylor expansion of the exponential function, we have
\begin{equation*}
\exp\{x\}-1=\sum_{i=1}^\infty \frac{1}{i!}x^i,
\end{equation*}
which implies
\begin{equation*}
\frac{\exp\{x\}-1}{x}=\sum_{i=1}^\infty \frac{1}{i!}x^{i-1}=\sum_{i=0}^\infty\frac{1}{(i+1)!}x^i.
\end{equation*}
Based on the Taylor expansion of the exponential function again, $\exp\{\frac{x}{2}\}$ can be written as
\begin{equation*}
\exp\{\frac{x}{2}\}=\sum_{i=0}^\infty \frac{1}{i!2^i}x^i
\end{equation*}
Define $g_1(i)=\frac{1}{(i+1)!}$ and $g_2(i)=\frac{1}{i!2^i}$. It is easy to show that $g_1(0)=g_2(0)=1$ and $g_1(1)=g_2(1)=\frac{1}{2}$.
For $i\ge 3$, we have
\begin{equation*}
g_1(i)-g_2(i)=\frac{2^i-(i+1)}{(i+1)!2^i}>0.
\end{equation*}
Since $x>0$, each term in $\frac{\exp\{x\}-1}{x}$ is no smaller than $\exp\{\frac{x}{2}\}$ and we reach the conclusion.\hfill$\Box$

\subsection*{Proof for Theorem \ref{theorem_h_convexity}}

{\bf Proof}. According to the requirement, $h'(\cdot)$ is monotonically increasing on $[0,\infty)$, implying that its second-order derivative $h''(\cdot)$ is positive, which is equivalent to the strong convexity of $h(\cdot)$ on $[0,\infty)$. $h(\cdot)$ is already required to be monotonically increasing. As both $h(\cdot)$ and $h'(\cdot)$ are required to be nonnegative, we only require that $h(0)\ge 0$ and $h'(0)\ge 0$ due to their monotonically increasing property.\hfill$\Box$

\subsection*{Proof for Theorem \ref{theorem_problem_convexity}}

{\bf Proof}: According to the scalar composition rule in Eq. (3.10) of \cite{bv04}, when $L(\mathcal{D}_i;\bm{\Theta})$ is convex with respect to $\bm{\Theta}$ and $h(\cdot)$ is convex and monotonically increasing, $h(L(\mathcal{D}_i;\bm{\Theta}))$ is convex with respect to $\bm{\Theta}$ and so is $\sum_{i=1}^m h(L(\mathcal{D}_i;\bm{\Theta}))$, leading to the validity of the first part. If further $r(\bm{\Theta})$ is convex with respect to $\bm{\Theta}$, both terms in the objective function of problem (\ref{obj_BMTL}) is convex with respect to $\bm{\Theta}$, making the whole problem convex. \hfill$\Box$

\subsection*{Proof for Theorem \ref{theorem_general_bound}}

{\bf Proof}. Since $h(\cdot)$ is a convex function, we can get
\begin{eqnarray*}
R_h(\bm{\Theta})&=&\frac{1}{m}\sum_{i=1}^mh(\mathbb{E}_{\mathcal{D}_i\sim\mu_i}[\hat{R}_i(\bm{\Theta})])\\
             &\le&\frac{1}{m}\sum_{i=1}^m\mathbb{E}_{\mathcal{D}_i\sim\mu_i}[h(\hat{R}_i(\bm{\Theta}))]\\
             &=&\mathbb{E}[\hat{R}_h(\bm{\Theta})]\\
             &\le& \hat{R}_h(\bm{\Theta})+\sup_{\mathbf{W}\in\mathcal{C}}\left\{\mathbb{E}[\hat{R}_h(\bm{\Theta})]-\hat{R}_h(\bm{\Theta})\right\}.
\end{eqnarray*}
When each pair of the training data $(\mathbf{x}^i_j,y^i_j)$ changes, the random variable $\sup_{\bm{\Theta}\in\mathcal{C}}\{\mathbb{E}[\hat{R}_h(\bm{\Theta})]-\hat{R}_h(\bm{\Theta})\}$ can change by no more than $\eta=\frac{2}{m}\exp\{\frac{1}{T}\}(\exp\{\frac{1}{n_0T}\}-1)$ due to the boundedness of the loss function $l(\cdot,\cdot)$. Then by McDiarmid's inequality \citep{mcdiarmid89}, we can get
{\scriptsize
\begin{align*}
&P\left(\sup_{\bm{\Theta}\in\mathcal{C}}\left\{\mathbb{E}[\hat{R}_h(\bm{\Theta})]-\hat{R}_h(\bm{\Theta})\right\}
-\mathbb{E}\left[\sup_{\bm{\Theta}\in\mathcal{C}}\left\{\mathbb{E}[\hat{R}_h(\bm{\Theta})]-\hat{R}_h(\bm{\Theta})\right\}\right]\ge t\right)\\
&\le \exp\left\{-\frac{2t^2}{mn_0\eta^2}\right\},
\end{align*}
}\noindent
where $P(\cdot)$ denotes the probability, and this inequality implies that with probability at least $1-\delta$,
{\small
\begin{align*}
\sup_{\bm{\Theta}\in\mathcal{C}}\left\{\mathbb{E}[\hat{R}_h(\bm{\Theta})]-\hat{R}_h(\bm{\Theta})\right\}
\le&\mathbb{E}\left[\sup_{\bm{\Theta}\in\mathcal{C}}\{\mathbb{E}[\hat{R}_h(\bm{\Theta})]-\hat{R}_h(\bm{\Theta})\}\right]\\
&+\sqrt{\frac{\eta^2mn_0}{2}\ln\frac{1}{\delta}}.
\end{align*}
}\noindent
If we have another training set $\{(\tilde{\mathbf{x}}^i_j,\tilde{y}^i_j)\}$ with the same distribution as $\{(\mathbf{x}^i_j,y^i_j)\}$, then we can bound $\mathbb{E}\left[\sup_{\bm{\Theta}\in\mathcal{C}}\{\mathbb{E}[\hat{R}_h(\bm{\Theta})]-\hat{R}_h(\bm{\Theta})\}\right]$ as
{\small
\begin{align*}
&\mathbb{E}\left[\sup_{\bm{\Theta}\in\mathcal{C}}\left\{\mathbb{E}[\hat{R}_h(\bm{\Theta})]-\hat{R}_h(\bm{\Theta})\right\}\right]\\
=&\mathbb{E}\left[\sup_{\bm{\Theta}\in\mathcal{C}}\left\{\mathbb{E}\left[\frac{1}{m}\sum_{i=1}^m h\left(\frac{1}{n_0}\sum_{j=1}^{n_0} l(f_i(\tilde{\mathbf{x}}^i_j),\tilde{y}^i_j)\right)\right]-\hat{R}_h(\bm{\Theta})\right\}\right]\\
\le& \mathbb{E}\left[\sup_{\bm{\Theta}\in\mathcal{C}}\left\{\frac{1}{m}\sum_{i=1}^m h\left(\frac{1}{n_0}\sum_{j=1}^{n_0} l(f_i(\tilde{\mathbf{x}}^i_j),\tilde{y}^i_j)\right)-\hat{R}_h(\bm{\Theta})\right\}\right].
\end{align*}
}\noindent
Multiplying the term by $m$ Rademacher variables $\{\sigma_i\}_{i=1}^m$, each of which is an uniform $\{\pm 1\}$-valued random variable, will not change the expectation since $\mathbb{E}[\sigma]=0$. Furthermore, negating a Rademacher variable does not change its distribution. So we have
{\scriptsize
\begin{align*}
&\mathbb{E}\left[\sup_{\bm{\Theta}\in\mathcal{C}}\left\{\frac{1}{m}\sum_{i=1}^m h\left(\frac{1}{n_0}\sum_{j=1}^{n_0} l(f_i(\tilde{\mathbf{x}}^i_j),\tilde{y}^i_j)\right)-\hat{R}_h(\bm{\Theta})\right\}\right]\\
=&\mathbb{E}\left[\sup_{\bm{\Theta}\in\mathcal{C}}\left\{\sum_{i=1}^m\frac{\sigma_i}{m} h\left(\frac{1}{n_0}\sum_{j=1}^{n_0} l(f_i(\tilde{\mathbf{x}}^i_j),\tilde{y}^i_j)\right)-\sum_{i=1}^m\frac{\sigma_i}{m}h(\hat{R}_i(\bm{\Theta}))\right\}\right]\\
\le&\mathbb{E}\left[\sup_{\bm{\Theta}\in\mathcal{C}}\left\{\sum_{i=1}^m\frac{\sigma_i}{m} h\left(\frac{1}{n_0}\sum_{j=1}^{n_0} l(f_i(\tilde{\mathbf{x}}^i_j),\tilde{y}^i_j)\right)\right\}\right]\\
&+\mathbb{E}\left[\sup_{\bm{\Theta}\in\mathcal{C}}\left\{\sum_{i=1}^m\frac{-\sigma_i}{m}h(\hat{R}_i(\bm{\Theta}))\right\}\right]\\
=&2\mathbb{E}\left[\sup_{\bm{\Theta}\in\mathcal{C}}\left\{\sum_{i=1}^m\frac{\sigma_i}{m}h(\hat{R}_i(\bm{\Theta}))\right\}\right].
\end{align*}
}\noindent
Note that $h(x)=\exp\{\frac{x}{T}\}$ is $\nu$-Lipschitz at $[0,1]$ where $\nu=\frac{1}{T}\exp{\frac{1}{T}}$. Due to the monotonic property of the loss function such as the cross-entropy loss and the hinge loss with respect to the first input argument, $\hat{R}_i(\bm{\Theta})$ is also a $\rho$-Lipschitz function. Then based on properties of the Rademacher compliexity, we can get
\begin{align*}
&2\mathbb{E}\left[\sup_{\bm{\Theta}\in\mathcal{C}}\left\{\sum_{i=1}^m\frac{\sigma_i}{m}h(\hat{R}_i(\bm{\Theta}))\right\}\right]\\
\le&4\nu\mathbb{E}\left[\sup_{\bm{\Theta}\in\mathcal{C}}\left\{\sum_{i=1}^m\frac{\sigma_i}{m}\hat{R}_i(\bm{\Theta})\right\}\right]\\
\le&8\rho\nu\mathbb{E}\left[\sup_{\bm{\Theta}\in\mathcal{C}}\left\{\sum_{i=1}^m\frac{\sigma_i}{n_0m}
\sum_{j=1}^{n_0}f_i(\mathbf{x}^i_j;\bm{\Theta})\right\}\right].
\end{align*}
Then by combining the above inequalities, we can reach the conclusion. \hfill$\Box$

\subsection*{Proof for Theorem \ref{theorem_bound_identity_function}}

{\bf Proof}. Since $h(\cdot)$ is convex, we can get
\begin{equation*}
h(R_I(\bm{\Theta}))\le R_h(\bm{\Theta}).
\end{equation*}
So we have
\begin{equation*}
R_I(\bm{\Theta})\le T\ln(R_h(\bm{\Theta})).
\end{equation*}
Then based on Theorem \ref{theorem_general_bound}, we reach the conclusion. \hfill$\Box$

\subsection*{Generalization Bound for Linear Models}

Based on Theorem \ref{theorem_general_bound}, we can analyze specific learning models. Here we consider a linear model where $\bm{\Theta}$ is a matrix with $m$ columns $\{\bm{\theta}_i\}$ each of which defines a learning function for a task as $f_i(\mathbf{x})=\langle\bm{\theta}_i,\mathbf{x}\rangle$. Here $r(\bm{\Theta})$ is defined as $r(\bm{\Theta})=\|\bm{\Theta}\|_F$ where $\|\cdot\|_F$ denotes the Frobenius norm. For problem (\ref{obj_BMTL_2}) with such a linear model, we have the following result.

\begin{theorem}\label{theorem_bound_linear_model}
When $\mathcal{C}=\{\bm{\Theta}|\|\bm{\Theta}\|_F\le \beta\}$, with probability at least $1-\delta$ where $\delta>0$, we have
\begin{align*}
R_h(\bm{\Theta})\le&\hat{R}_h(\bm{\Theta})+\frac{8\rho\nu\beta}{\sqrt{m}}+\sqrt{\frac{\eta^2mn_0}{2}\ln\frac{1}{\delta}}.
\end{align*}
\end{theorem}
{\bf Proof.} According to Theorem \ref{theorem_general_bound}, we only need to upper-bound $\mathbb{E}\left[\sup_{\bm{\Theta}\in\mathcal{C}}\left\{\sum_{i=1}^m\frac{\sigma_i}{n_0m}
\sum_{j=1}^{n_0}f_i(\mathbf{x}^i_j;\bm{\Theta})\right\}\right]$. By defining $\bar{\mathbf{x}}^i=\frac{1}{n_0}\sum_{j=1}^{n_0}\mathbf{x}^i_j$, we have
{\small
\begin{align*}
&\mathbb{E}\left[\sup_{\bm{\Theta}\in\mathcal{C}}\left\{\sum_{i=1}^m\frac{\sigma_i}{n_0m}
\sum_{j=1}^{n_0}f_i(\mathbf{x}^i_j;\bm{\Theta})\right\}\right]\\
=&\frac{1}{m}\mathbb{E}\left[\sup_{\bm{\Theta}\in\mathcal{C}}\left\{\sum_{i=1}^m\sigma_i\langle\bm{\theta}_i,\bar{\mathbf{x}}^i\rangle
\right\}\right]\\
=&\frac{1}{m}\mathbb{E}\left[\sup_{\bm{\Theta}\in\mathcal{C}}\left\{\langle\bm{\Theta},
[\sigma_1\bar{\mathbf{x}}^1,\ldots,\sigma_m\bar{\mathbf{x}}^m]\rangle
\right\}\right]\\
\le&\frac{1}{m}\mathbb{E}\left[\sup_{\bm{\Theta}\in\mathcal{C}}\left\{\|\bm{\Theta}\|_F\|
[\sigma_1\bar{\mathbf{x}}^1,\ldots,\sigma_m\bar{\mathbf{x}}^m]\|_F
\right\}\right]\\
\le&\frac{\beta}{m}\mathbb{E}\left[
\|[\sigma_1\bar{\mathbf{x}}^1,\ldots,\sigma_m\bar{\mathbf{x}}^m]\|_F\right]\\
\le&\frac{\beta}{m}\sqrt{\mathbb{E}\left[
\|[\sigma_1\bar{\mathbf{x}}^1,\ldots,\sigma_m\bar{\mathbf{x}}^m]\|^2_F\right]}\\
=&\frac{\beta}{m}\sqrt{\mathbb{E}\left[
\sum_{i=1}^m\sigma_i^2\|\bar{\mathbf{x}}^i\|_2^2\right]}\\
=&\frac{\beta}{m}\sqrt{\mathbb{E}\left[
\sum_{i=1}^m\|\bar{\mathbf{x}}^i\|_2^2\right]}\\
\le&\frac{\beta}{\sqrt{m}},
\end{align*}
}\noindent
where the first inequality is due to the Cauchy-Schwartz inequality, the second inequality holds because of the constraint on $\bm{\Theta}$, the third inequality is due to the Jensen's inequality based on the square root function, the fourth inequality holds since the $\ell_2$ norm of each data point is upper-bounded by 1 and $\bar{\mathbf{x}}^i$ is the average of data points in the $i$th task. \hfill$\Box$

\end{document}